%% file: main.tex
\documentclass{article}

\usepackage[preprint]{neurips_2026}

\usepackage[T1]{fontenc}
\usepackage{tikz}
\usepackage[pagebackref=true,colorlinks,allcolors=blue]{hyperref}
\usepackage{url}            
\usepackage{booktabs}       
\usepackage{amsfonts}       
\usepackage{nicefrac}       
\usepackage{microtype}      
\usepackage{xcolor}         
\usepackage{times}
\usepackage{epsfig}
\usepackage{graphicx}
\usepackage{amsmath}
\usepackage{amssymb}
\usepackage{amsthm}
\usepackage{paralist}
\usepackage{pifont}
\usepackage{multirow}
\usepackage{bm}
\usepackage{bbm}
\usepackage{diagbox}
\usepackage{makecell}
\usepackage{float}
\usepackage{stfloats}
\usepackage{algorithm}  
\usepackage{color, colortbl}
\usepackage{arydshln}
\usepackage{marvosym}
\usepackage{booktabs}
\usepackage{longtable}
\usepackage{subcaption}
\usetikzlibrary{arrows.meta,calc,fit,positioning}
\usepackage{algpseudocode}

\theoremstyle{definition}

\theoremstyle{plain}

\renewcommand{\eqref}[1]{Eq.~(\ref{#1})}

\newcommand{\gr}{\rowcolor[gray]{.95}} 

\usepackage{listings}

\lstdefinelanguage{yaml}{
keywords={name,hook_point,summary,capabilities,activation_policy,signal,
  default_mode,default_threshold,planning_phases,inputs,required_cache_fields,
  required_meta_fields,required_capability_fields,hook_capability_fields,
  outputs,runtime_contract,execution_class,affects_current_frame,latest_wins},
sensitive=false,
comment=[l]{\#},
morestring=[b]",
morestring=[b]'
}

\lstdefinestyle{profileyaml}{
language=yaml,
basicstyle=\ttfamily\scriptsize,
keywordstyle=\bfseries,
commentstyle=\color{gray},
frame=single,
rulecolor=\color{black!25},
showstringspaces=false,
breaklines=true,
breakatwhitespace=false,
columns=fullflexible,
xleftmargin=0.6em,
framexleftmargin=0.6em
}

\title{DrivingAgent: Design and Scheduling Agents \\ for Autonomous Driving Systems}

\author{
Zhongyu Xia$^1$\quad Wenhao Chen$^1$\quad Yongtao Wang$^1$\thanks{Corresponding author.}\quad Ming-Hsuan Yang$^2$
\\
$^1$Wangxuan Institute of Computer Technology, Peking University
\\
$^2$University of California, Merced
\\
\texttt{\{xiazhongyu,wyt\}@pku.edu.cn \quad wenhaochen@stu.pku.edu.cn}
\\
\texttt{mhyang@ucmerced.edu}
}

\begin{document}

\maketitle
\input{sections/0_abstract}
\input{sections/1_intro}
\input{sections/2_related_work}

\input{sections/3_method}
\input{sections/4_experiments}
\input{sections/6_conclusion}


\newpage
\small
\bibliographystyle{plain}
\bibliography{references}


\clearpage
\appendix
\input{sections/7_appendix}



\end{document}

%% file: sections/0_abstract.tex
\begin{abstract}


Many autonomous driving systems are increasingly incorporating foundation models to improve generalization and handle long-tail scenarios. However, this trend introduces two key challenges: (i) the manual and labor-intensive process of designing and integrating new models, and (ii) the lack of intelligent, dynamic scheduling mechanisms to meet strict real-time constraints.
While Large Language Model (LLM)-based agents offer a promising avenue for automation, existing frameworks are ill-suited for autonomous driving. Specifically, they fail to distinguish between the fundamentally different requirements of system design and real-time scheduling, treat modules as opaque black boxes, and are not designed for continuous operation.
To address these limitations, we propose \textbf{DrivingAgent}, a novel agent framework tailored to the dual challenges of autonomous driving system design and scheduling. In the design phase, DrivingAgent automates module development by interpreting system architecture, generating code, and validating modules via super-network training. In the scheduling phase, it employs a lightweight LLM trained with reinforcement learning to dynamically orchestrate system modules in real time, supported by a structured memory that integrates long-term storage with timestamped short-term context.
Experimental results demonstrate that DrivingAgent achieves a superior speed--accuracy trade-off on both the nuScenes and Bench2Drive benchmarks.

\end{abstract}

%% file: sections/1_intro.tex
\section{Introduction}
\label{sec:intro}

Autonomous driving is a core application of deep learning. An autonomous driving system takes as input sensor data, navigation signals, or natural language commands, and predicts either future trajectories or control signals for the ego vehicle.
A wide range of paradigms has been proposed for autonomous driving systems. However, to improve generalization, handle long-tail scenarios, and satisfy legal and ethical constraints, these systems are becoming increasingly large and complex. Some approaches are built on Vision-Language Models (VLMs) or video generation models, in which vision-centric representations are augmented with action experts to form Vision-Language-Action (VLA) models~\cite{fu2025orion, renz2025simlingo} or World Action Models (WAMs)~\cite{han2025percept}. These frameworks are further extended with reasoning capabilities, such as Mixture-of-Thoughts (MoT)~\cite{li2026unidrivevla}.
Other approaches rely on multimodal fusion of vision and point clouds. They constrain 3D representations using perception and prediction objectives, while employing dedicated planning modules for decision-making~\cite{hu2023uniad, jiang2023vad, zheng2024genad, henetpp}. To better address long-tail scenarios and ensure compliance with legal and ethical requirements, additional components such as VLMs and knowledge graphs are increasingly integrated~\cite{xia2025knowval}.
These developments introduce two key challenges.

\begin{figure}[htb]
\centering
\includegraphics[width=\textwidth]{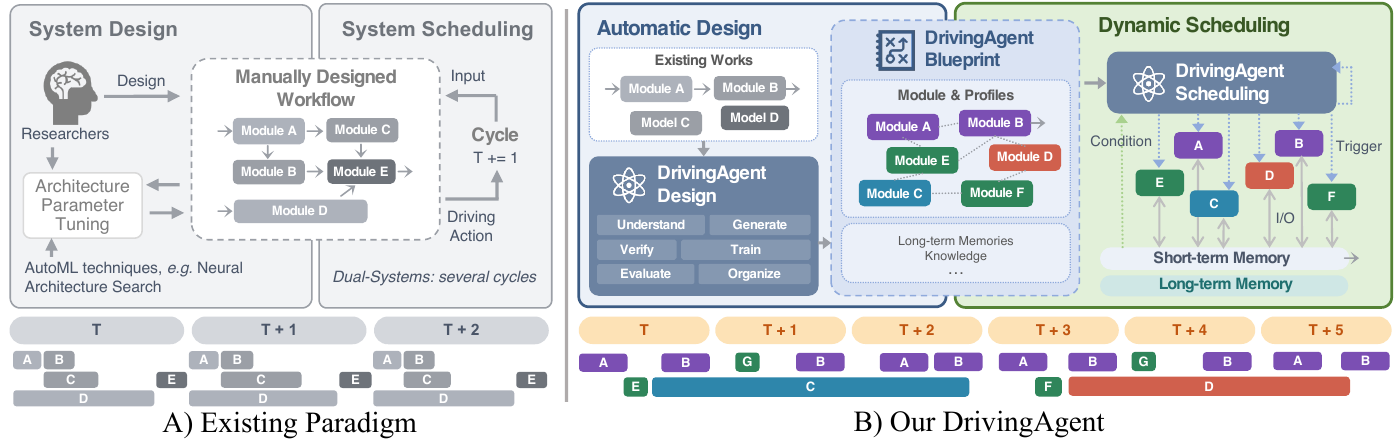}
\caption{Compared with existing autonomous driving paradigms, DrivingAgent automatically integrates prior methods and designs, trains, and validates new adapter modules. During inference, it leverages a design blueprint and a lightweight LLM trained via reinforcement learning, conditioned on a timestamped memory bank, to perform dynamic concurrent scheduling, thereby improving the speed-accuracy trade-off.}
\label{fig:intro}
\vspace{-5pt}
\end{figure}

First, incorporating a new prior module typically requires substantial manual effort in network design and training for system integration. For example, prior work designs specialized adapters~\cite{lin2025vl} to extract visual prompts from the attention maps of a Vision-Language Model (VLM). A framework that can \textbf{automatically design such modules at scale and perform automated validation and selection} would substantially reduce the research and development burden.

Second, autonomous driving imposes stringent real-time constraints. Many modules in existing systems do not need to be invoked at high frequency, such as natural language encoders or knowledge graph retrieval components. While some prior work adopts a dual-system design~\cite{tian2024drivevlm} that executes slower modules (e.g., VLMs) at reduced frame rates, these approaches remain suboptimal because they lack \textbf{dynamic, context-aware scheduling}.
For example, when no new perceptual elements are observed, and the query remains unchanged, repeatedly invoking the knowledge retrieval module is unnecessary. Similarly, mapping modules can operate at different update frequencies depending on the driving context, such as high-speed motion versus a stationary waiting state.

With the rapid advancement of Large Language Models (LLMs), numerous agent frameworks have been proposed~\cite{yao2022react,schick2023toolformer,park2023generative,shinn2023reflexion,wang2023voyager,wu2024autogen,hong2023metagpt}. Such agents can autonomously interpret goals, devise plans, invoke tools, and execute tasks, making them a natural fit for addressing the aforementioned challenges. However, existing agent frameworks are not directly applicable to the design and scheduling problems in autonomous driving for several reasons.
\textbf{First, design and scheduling are tightly coupled yet fundamentally distinct tasks.} A design agent typically relies on tools such as debuggers and syntax checkers, operating across both training and inference stages, where latency is less critical. In contrast, a scheduling agent interacts with system modules as executable components, with primary emphasis on real-time inference performance.
\textbf{Second, existing agent frameworks often treat tools as black boxes.} In autonomous driving, however, effective design and scheduling of neural networks require explicit understanding of model internals and structural dependencies.
\textbf{Third, current agent frameworks are primarily designed for one-off task completion rather than continuous operation.} As a result, they lack mechanisms to reason about runtime duration and information staleness, and do not provide dedicated designs for both long-term and short-term memory in persistent, real-time environments.

Therefore, we propose \textbf{DrivingAgent}, a unique agent framework that integrates automated module design with intelligent real-time scheduling to address the intertwined challenges of building and orchestrating autonomous driving systems.
Our contributions include:
%
%
\textbf{(1)} We introduce the DrivingAgent framework, whose design phase analyzes existing neural network modules, constructs a structured design blueprint, and performs automated code generation followed by super-network training for validation. Each accepted generated module is exported as an agent-tool profile.
%
%
\textbf{(2)} DrivingAgent features a scheduling phase that employs a lightweight LLM, fine-tuned via reinforcement learning with a carefully designed hybrid reward, to dynamically orchestrate neural network modules in real time. The scheduler leverages both long-term memory and timestamped short-term memory to achieve context-aware module coordination.
\textbf{(3)} 
To our knowledge, DrivingAgent is the first to systematically explore automated design and dynamic scheduling for autonomous driving, and the first to address this with an LLM agent framework.
DrivingAgent achieves a superior speed-accuracy trade-off on both the nuScenes and Bench2Drive benchmarks.

%% file: sections/2_related_work.tex
\section{Related Work}
\label{sec:related}

\noindent \textbf{Autonomous Driving Systems.}
Autonomous driving systems span diverse paradigms, including modular pipelines, end-to-end models, VLAs, and WAMs.
Within the end-to-end paradigm, ST-P3~\cite{hu2022st} unifies perception, prediction, and planning, and UniAD~\cite{hu2023uniad} extends this with BEV representations and task-specific decoders.
VAD~\cite{jiang2023vad} introduces vectorized scene representations, while PARA-Drive~\cite{weng2024drive} studies structured multi-task supervision.
GenAD~\cite{zheng2024genad} and SparseDrive~\cite{sun2025sparsedrive} focus on sparse feature representations to improve planning and reduce error accumulation.
Recent works emphasize temporal consistency and diversity, including MomAD~\cite{song2025don}, BridgeAD~\cite{zhang2025bridging}, DiffusionDrive~\cite{liao2025diffusiondrive}, and HENet++~\cite{henetpp}.
Additional designs improve planning and efficiency: TCP~\cite{wu2022trajectory}, ThinkTwice~\cite{jia2023think}, DriveAdapter~\cite{jia2023driveadapter}, and DriveTransformer~\cite{jia2025drivetransformer}.
In the VLA paradigm, DriveLM~\citep{sima2024drivelm} formulates reasoning as graph-based VQA, while ORION~\citep{fu2025orion}, SimLingo~\citep{renz2025simlingo}, and KnowVal~\citep{xia2025knowval} integrate language, control, and retrieval.
Overall, improved generalization and long-tail robustness come at the cost of increased architectural complexity and computational demands.

Some methods attempt a simple parallel design, running the VLM and the planner cyclically at different frequencies.
DriveVLM-Dual~\citep{tian2024drivevlm} pairs a low-frequency VLM with a real-time classical planner and asynchronously injects VLM trajectories.
AsyncDriver~\citep{chen2024asynchronous} injects LLM scene-instruction features into a transformer planner at controllable intervals.
SOLVE~\citep{chen2025solve} temporally decouples a VLM and an end-to-end branch over a shared visual encoder. 
However, the scheduling mechanisms of these methods remain relatively simple, requiring manually designed interfaces and lacking the ability to perform dynamic scheduling according to the scenario. 
For example, even when the language input remains completely unchanged, any of the aforementioned systems will still repeatedly run the language encoder, resulting in redundant computation.

\noindent \textbf{Agent Frameworks.}
LLM-based agent frameworks can be broadly categorized into three paradigms: reasoning-and-acting loops, multi-agent collaboration, and tool-use augmentation. ReAct~\citep{yao2022react} pioneers the interleaving of reasoning and action within a unified loop. Reflexion~\citep{shinn2023reflexion} and Voyager~\citep{wang2023voyager} extend this paradigm by introducing verbal self-improvement and open-ended skill acquisition, respectively.
In the multi-agent setting, AutoGen~\citep{wu2024autogen} and MetaGPT~\citep{hong2023metagpt} organize LLMs into role-based collaborative structures, while Generative Agents~\citep{park2023generative} incorporate memory and reflection mechanisms to simulate realistic behaviors. For tool use, Toolformer~\citep{schick2023toolformer} enables models to learn to invoke APIs from unlabeled data.
Despite these advances, existing frameworks are not well-suited to autonomous driving. Their tools are typically treated as opaque API calls, rather than neural modules with explicit tensor I/O and compositional interfaces. They are designed for one-shot objectives, rather than persistent, event-driven environments with cache validity and information expiration. Moreover, they lack a principled separation between design-time authoring and inference-time scheduling, a distinction that is critical for autonomous driving systems.
DriveAgent-R1~\cite{zhengdriveagent} is the first to apply an LLM agent to autonomous driving scenarios, treating vision input acquisition, cropping, depth estimation, and perception as agent tools. However, it is limited to scene understanding and cannot perform planning tasks. Moreover, it does not address design and scheduling problems, and its latency ranges from 6740 ms to 7910 ms, making it incapable of real-time operation.

%% file: sections/3_method.tex
\section{Method}
\label{sec:method}

As shown in Figure~\ref{fig:overview}, DrivingAgent addresses two coupled operations in a heterogeneous driving stack: (i) generating new neural agent tools, and (ii) deciding when these tools should be invoked online. The \emph{Design Agent} (Section~\ref{sec:method:design}) parses the base planner’s white-box interface and translates targeted capability gaps into validated agent tools. The \emph{Concurrent Scheduling Agent} (Section~\ref{sec:method:edge}) operates in an event-time runtime, consuming the resulting contracts to control tool invocation, reuse, asynchronous updates, and temporary deactivation based on runtime evidence. The input frame rate is fixed by the sensor stream; the scheduler modulates the effective compute FPS by adjusting the amount of agent-tool computation requested.

\subsection{Design Agent for White-Box Neural Networks}
\label{sec:method:design}
\begin{figure}[tp]
\centering
\includegraphics[width=\linewidth]{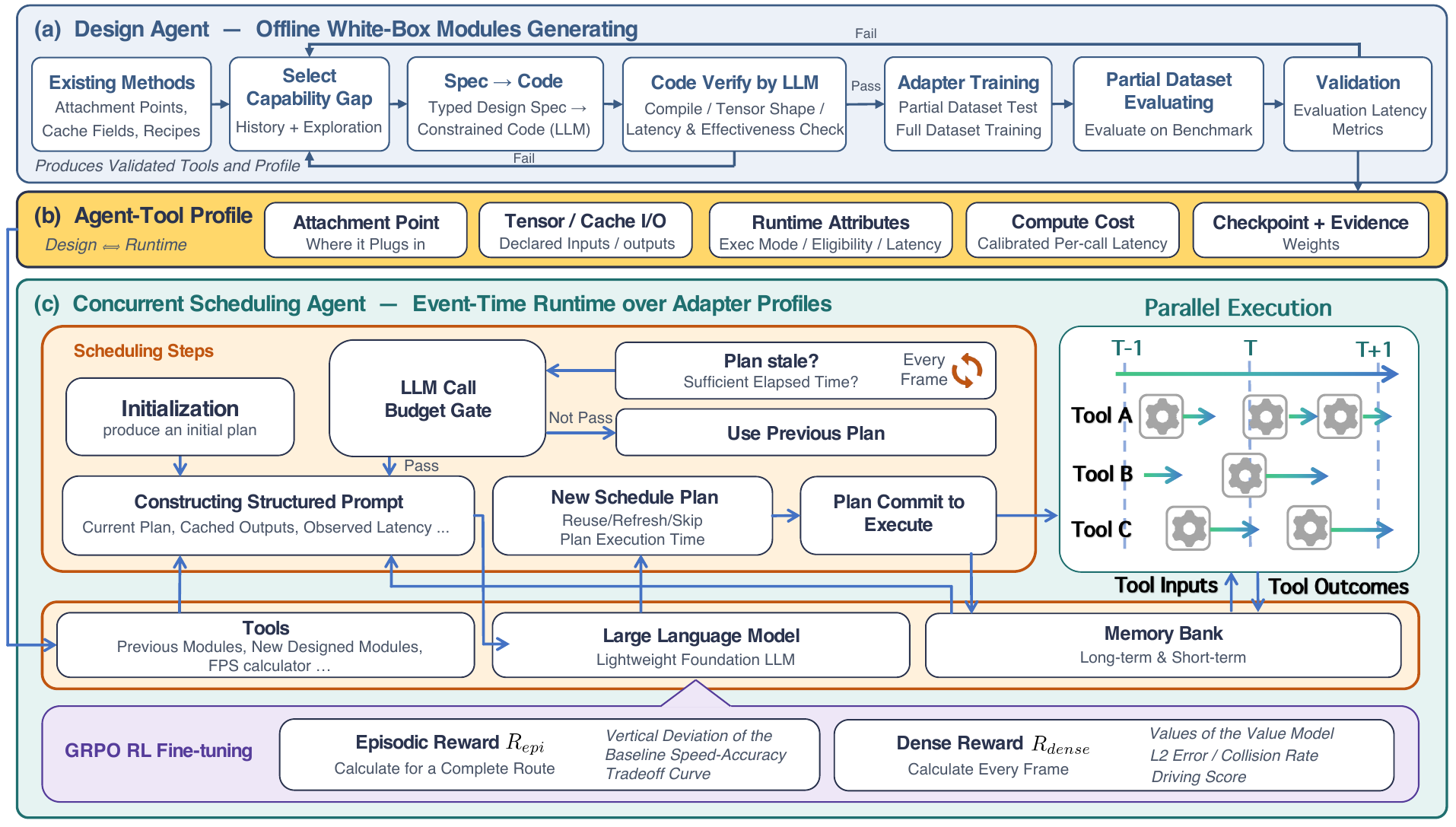}
\caption{\textbf{DrivingAgent overview.}
DrivingAgent separates offline adapter construction from online scheduling through a structured \emph{Tool Profile}.
(a) The \emph{Design Agent} discovers planner capability gaps, generates and verifies adapter code, trains candidate modules, and validates them before deployment.
(b) The \emph{Agent-Tool Profile} serves as the design--runtime contract, encoding attachment points, tensor/cache interfaces, runtime attributes, compute cost, and checkpoint evidence.
(c) The \emph{Scheduling Agent} uses these profiles to decide whether to reuse or refresh schedules, execute tools in parallel, and update memory from runtime outcomes. Episodic and dense rewards provide reinforcement signals for improving scheduling decisions.
}
\label{fig:overview}
\end{figure}

The Design Agent is the offline part of DrivingAgent. As shown in Figure~\ref{fig:overview}(a), its input is a pool of \emph{Existing Methods}: a white-box planner with exposed internal interfaces, together with the available external components that have not been attached to the planner. It converts this pool into a design blueprint, analyzes the capability gap, uses the LLM to generate a typed specification and code, verifies the generated code, trains and evaluates the candidate, and admits it only if the validation gate passes. Its outputs are validated modules and an agent-tool profile for the Scheduling Agent.

\textbf{Architecture comprehension and gap selection.}
The Design Agent operates offline, transforming module generation into an interface-grounded process for downstream scheduling. We use \emph{agent tool} to denote a module produced by the Design Agent and exposed to the Scheduling Agent as a tool, such as a learned adapter or an external perception or retrieval component. The agent first maps the existing methods into a finite, typed design space. An \emph{attachment point} defines a boundary where an agent tool can read intermediate states or publish reusable outputs. A \emph{cache field} is a named tensor or semantic value whose validity can be managed at runtime. A \emph{supervision recipe} is a registered training template that resolves a declared capability into concrete targets, losses, and weighting schemes. These abstractions enable reasoning over neural-network internals without treating the planner as an opaque end-to-end function.

The agent then identifies a \emph{capability gap}, defined as an interface-grounded function that is insufficiently addressed by the current planner or deployed agent tools. Gap selection integrates the typed interface, existing agent tools, design history, and runtime feedback. Rather than enumerating arbitrary modules, the agent selects targets whose outputs can be consistently attached, trained, and scheduled under a unified runtime contract.

\textbf{Spec-first module design.}
Given a selected gap, the LLM generator first produces a \emph{design specification} rather than free-form code. The specification binds the candidate to an attachment point, defines required inputs and outputs, resolves supervision recipes, and records \emph{runtime attributes} such as current-frame eligibility, execution mode, and latency constraints. Code generation is then constrained to satisfy this specification. This decouples architectural reasoning from implementation: the design stage uses the white-box blueprint, while the online scheduler operates on a compact, well-defined runtime contract.

\textbf{Code verification and module validation.}
Before training, the generated module is checked against the design specification. An LLM-based verifier and deterministic checks ensure that declared inputs must be available at the attachment point, declared outputs must be produced, supervision targets must be well-defined, and runtime attributes must be explicit. Violations are returned as structured repair signals, enabling iterative refinement of the spec--code pair prior to training.
Candidates that pass verification proceed to staged module training: a lightweight probe first estimates trainability, quality impact, and latency overhead, while full training and benchmark evaluation are reserved for candidates that pass this initial screening. After full training and evaluation, the validation gate checks whether the candidate satisfies the required quality metrics and latency budget. Candidates that fail this gate are returned for repair, while candidates that pass are admitted as schedulable tools.

\textbf{Agent-tool profile handoff.}
A validated design specification, together with its training evidence and acceptance record, is compiled into an \emph{agent-tool profile}. This profile is the sole interface consumed by the scheduler: it specifies where the module attaches, its required inputs and produced outputs, the conditions under which its outputs are valid, and its associated compute cost. The full agent-tool profile schema and an example registry-to-runtime conversion are provided in the Appendix.

%

\subsection{Concurrent Scheduling Agent for Complex Intelligent Systems}
\label{sec:method:edge}
The concurrent Scheduling Agent addresses the online counterpart of the design problem. Given a base planner and a set of validated agent tools, it determines which tool outputs are eligible for the current frame, which tools should be refreshed for future frames, and which should be temporarily suppressed when recent stream-level evidence indicates low utility. The central design choice is to formulate this as an event-time control problem over explicit module contracts, rather than as per-frame LLM invocation or a fixed low-frequency auxiliary pipeline.

We adopt \emph{event time} as the source timestamp of the driving stream, in contrast to wall-clock arrival or replay time. This distinction ensures that cache validity, plan activation, suppression windows, and asynchronous refresh remain causally aligned with the driving sequence, even when replay or background execution proceeds slower than real time.
\subsubsection{Agent Framework}
The Scheduling Agent operates as an event-time scheduler over agent tools. It maintains a timescale memory bank that records schedule plans, tool outcomes, and latency feedback. When the active plan becomes stale, a lightweight LLM generates an updated plan for upcoming frames, specifying which tools to activate and whether to recompute or reuse cached outputs. At each frame, deterministic routing resolves the plan against the available tools and current memory state, while tool execution proceeds asynchronously, updating the memory bank without blocking the control path. 

\textbf{Abstraction for agent tools.}
The scheduler does not access agent tool implementations; instead, it operates on agent-tool profiles. Each profile encodes only scheduling-relevant information, including attachment location, required inputs, updatable cached values, default execution mode, latency budget, and a concise description of capability. And we use the \emph{fixed latency table} to record the theoretical latency of a single forward call for each existing module, excluding data loading; its entries are detailed in the Appendix.

\textbf{FPS calculator for feedback.}
The scheduler employs an FPS estimator to predict throughput based on the selected execution modes and the fixed latency table. After execution, the runtime measures the achieved throughput. The predicted FPS, measured FPS, and their discrepancy are stored in short-term memory, enabling subsequent scheduling decisions to correct cost estimation errors. 

\textbf{Timescale event memory bank.}
The scheduler maintains memory at two temporal scales. Long-term memory stores cross-frame state, including the active plan, pending updates, module completion records, and reusable cached outputs. Short-term memory provides a transient context for each LLM call, capturing observable state such as ego motion, background tool execution, runtime latency, and recent outcomes. Initialized empty, memory evolves over time as evidence accumulates.

\textbf{Schedule plan initialization.}
At the start of a stream, only agent-tool profiles are available. The scheduler queries the LLM to produce an initial plan, which becomes the active schedule. Subsequent updates are triggered by event-time evidence rather than unconditional LLM invocation. 

\textbf{LLM interface.}
For each invocation, the scheduler constructs a structured prompt from agent-tool profiles and short-term context, including the current plan, cached outputs, background executions, and observed latency. The LLM returns a schedule plan specifying tool policies: \emph{Reuse} (use cached outputs), \emph{Refresh} (recompute for future frames), or \emph{Skip} (disable temporarily). Predicted throughput is computed, compared with measured performance, and written back to memory. 

\textbf{LLM call budget.}
The scheduler does not invoke the LLM at every frame. Instead, it updates plans only when they become stale and certain conditions are met, including sufficient elapsed time, absence of pending LLM calls, bounded background execution, and adherence to latency constraints. Otherwise, the current plan is retained. 

\textbf{Per-frame execution.}
At runtime, the plan is applied subject to resource and validity constraints. Reuse is permitted only when cached outputs remain valid, and refresh is triggered only when inputs are available and execution budgets allow. Refresh operations are dispatched asynchronously after the current frame, updating memory upon completion without blocking control decisions. If constraints are violated, actions degrade gracefully to reuse or skip. Consequently, the input frame rate remains fixed, while computational load varies dynamically across frames.

\subsubsection{Fine-tuning the Scheduling Agent with GRPO}
\label{sec:method:edge:fine-tune}

The zero-shot scheduler defines the runtime interface: it consumes a structured prompt built from agent-tool profiles and the memory bank, and outputs a schedule plan that affects future frames. GRPO fine-tuning preserves this interface and trains the policy to favor plans that improve the speed-accuracy trade-off while maintaining the same event-time semantics. Each sampled plan is rolled out over a short horizon, scored by quality and theoretical compute latency, and normalized against other plans from the same prompt. Fine-tuning thus reshapes the policy over valid plans.

\textbf{Reward design.}
Let $q$ denote the complete structured scheduler prompt at one call of the LLM. The trainable LLM reads this prompt and returns one schedule plan. We write the trainable LLM as $\pi_\theta$, where $\theta$ denotes its model weights.
In fine-tuning, we do not decode a single deterministic plan from $q$. Instead, we sample several possible schedule plans from the same prompt, compare their rewards, and train the LLM to prefer the better ones. We write one such sampled plan as $o\sim\pi_\theta(\cdot\mid q)$. The LLM output tokens for this plan are $o=(o_1,\ldots,o_L)$, where $L$ is the number of tokens and $o_s$ is the token at position $s$.
The Scheduling Agent executes plan $o$ for the next $H$ source frames. In the formulas below, $t\in\{1,\ldots,H\}$ means the $t$-th frame among these $H$ frames. For frame $t$, let $m_t$ be the reported planning metric and define $M_t$ so that larger is better: $M_t=-m_t$ for L2 error or collision rate, and $M_t=m_t$ for Driving Score. Let $\ell^{\mathrm{comp}}_t>0$ be the theoretical compute latency of frame $t$ in seconds, obtained from the fixed latency table used by the FPS calculator. We summarize the $H$ frames by mean quality and theoretical compute throughput:

\begin{equation}
M^{\star}(o)
  = \frac{1}{H}\sum_{t=1}^{H} M_t,
\qquad
F_{\mathrm{comp}}^{\star}(o)
  = \frac{H}{\sum_{t=1}^{H} \ell^{\mathrm{comp}}_t}.
\label{eq:segment-stats}
\end{equation}
Here $M^{\star}(o)$ is the mean signed quality of plan $o$, and $F_{\mathrm{comp}}^{\star}(o)$ is its theoretical compute FPS over the same $H$ frames.

\textit{Speed--accuracy episodic reward.}
The episodic term compares plan $o$ with a calibrated speed--accuracy reference line. We build this line from $B$ calibration points, denoted by $\mathcal{B}=\{(M_1,F_1),\ldots,(M_B,F_B)\}$. Each calibration point is one measured zero-shot scheduler result on the same base planner and agent tools. The subscript $b$ indexes one calibration point: $M_b$ is its mean signed quality and $F_b$ is its theoretical compute FPS. These zero-shot points use the same fixed latency table as $M^\star(o)$ and $F_{\mathrm{comp}}^\star(o)$. We choose two calibration points $a$ and $c$ as anchors, with $|F_c-F_a|>\epsilon_F$. Here $\epsilon_F>0$ is a small threshold that avoids using two anchors with nearly identical FPS. The line through these two anchors is
\begin{equation}
f(F)
= M_a + \frac{M_c-M_a}{F_c-F_a}\,(F-F_a).
\label{eq:chord}
\end{equation}
where $F$ is the FPS value at which the line is evaluated, and $f(F)$ is the reference signed quality at that FPS. We only use anchor pairs whose line is no lower than every calibration point, i.e., $M_b\le f(F_b)$ for all $b$; if multiple pairs satisfy this condition, we use the closest such line. Let $F_{\min}=\min(F_a,F_c)$ and $F_{\max}=\max(F_a,F_c)$. We clip the plan throughput to this calibrated FPS range:
\[
\bar F(o)=\min(\max(F_{\mathrm{comp}}^{\star}(o),F_{\min}),F_{\max}).
\]
Thus $\bar F(o)$ equals $F_{\min}$ if the plan is slower than the calibrated range, equals $F_{\max}$ if it is faster, and otherwise equals $F_{\mathrm{comp}}^{\star}(o)$. This avoids evaluating the reference line outside the calibrated FPS range; the unclipped $F_{\mathrm{comp}}^{\star}(o)$ is still reported in the results. The episodic reward is
\begin{equation}
R_{\mathrm{epi}}(o)
  = M^{\star}(o) - f\!\left(\bar F(o)\right).
\label{eq:r-epi}
\end{equation}
Thus $R_{\mathrm{epi}}(o)$ is positive when plan $o$ achieves a higher mean quality than the reference line at the same clipped FPS, and negative when it falls below that line. This term gives one scalar reward for the $H$-frame execution of plan $o$.

\textit{Dense per-frame reward.}
Let $V_\psi(t)$ be an optional frozen value model with fixed parameters $\psi$. It maps the committed trajectory and retrieved context of frame $t$ to a scalar quality score. If this scorer is not used, we set $\beta=0$. The per-frame reward and its $H$-frame average are
\begin{equation}
r^{\mathrm{dense}}_{t}=M_t+\beta\,V_\psi(t),
\qquad
R_{\mathrm{dense}}(o)=\frac{1}{H}\sum_{t=1}^{H} r^{\mathrm{dense}}_{t},
\qquad
\beta \ge 0.
\label{eq:r-dense}
\end{equation}
Here $r^{\mathrm{dense}}_{t}$ is the reward for frame $t$, $R_{\mathrm{dense}}(o)$ is the average dense reward for plan $o$, and $\beta$ controls how much the optional value scorer contributes.

\textit{Total reward.}
The final reward for plan $o$ is
\begin{equation}
R(o)=R_{\mathrm{epi}}(o)+\alpha\,R_{\mathrm{dense}}(o),
\qquad
\alpha \ge 0.
\label{eq:r-total}
\end{equation}
Here $\alpha$ controls the weight of the dense per-frame term.

\textbf{GRPO fine-tuning with future-frame deployment.}
The reward in Eq.~\ref{eq:r-total} is assigned to a complete schedule plan, while the scheduler generates that plan as an LLM output sequence. We use GRPO~\citep{shao2024deepseekmath} to update the LLM from these plan-level rewards without training a separate critic. For each prompt $q$, the old policy $\pi_{\theta_{\mathrm{old}}}$ samples a group of $G$ schema-valid plans,
\[
\{o_i\}_{i=1}^{G}\sim\pi_{\theta_{\mathrm{old}}}(\cdot\mid q),
\]
and each plan receives the reward $R_i=R(o_i)$. The advantage of plan $o_i$ is computed by normalizing its reward within this same group:
\begin{equation}
\widehat A_i =
\frac{R_i-\mu_R}{\sigma_R+\epsilon_\sigma},
\qquad
\mu_R=\frac{1}{G}\sum_{j=1}^{G}R_j,
\qquad
\sigma_R^2=\frac{1}{G}\sum_{j=1}^{G}(R_j-\mu_R)^2.
\label{eq:grpo-adv}
\end{equation}
Here $\mu_R$ and $\sigma_R$ are the mean and standard deviation of the $G$ rewards, and $\epsilon_\sigma$ is a small constant for numerical stability. Thus $\widehat A_i$ is positive when plan $o_i$ performs better than the other plans sampled from the same prompt, and negative when it performs worse. Because the reward is observed for the whole plan, the same $\widehat A_i$ is assigned to every token in the serialized output $o_i$. The policy $\pi_\theta$ is then updated with the clipped GRPO objective and a KL penalty to a frozen SFT reference policy $\pi_{\mathrm{ref}}$. This fine-tuning changes only the probabilities of schema-valid plans; the output schema, cache rules, and future-frame activation boundary remain the same as in the zero-shot scheduler.

%% file: sections/4_experiments.tex
\section{Experimental Results}
\label{sec:exp}

\subsection{Datasets and Metrics}
\label{sec:exp:data_metrics}

We evaluate DrivingAgent on two benchmarks: nuScenes~\citep{caesar2020nuscenes} and Bench2Drive~\citep{jia2024bench2drive}.
nuScenes contains $1$k scenes and $\sim$35k samples. We use it for open-loop planning and report L2 error between predicted and human trajectories, together with the collision rate. These metrics are calculated under the VAD planning metric protocol. 
Prior work~\citep{xia2025knowval} shows that the collision rate better reflects planning quality than L2, as trajectory similarity does not guarantee safety or rule compliance.
Bench2Drive evaluates closed-loop execution in simulation. Following the official protocol, we report Driving Score (route completion with infraction penalties) and Success Rate. It provides 10k training segments.

\subsection{Implementation Details}
\label{sec:exp:impl}

We use the Design Agent to assemble existing modules and generate new ones. The system integrates:
HENet++ (nuScenes only), decomposed into backbone, sparse 3D encoding and detection, dense 3D encoding and occupancy, and planner;
SimLingo~\citep{renz2025simlingo} (Bench2Drive only);
DiffusionDrive~\cite{liao2025diffusiondrive}, decomposed into backbone, perception, and planner (with dataset-specific weights);
VLM (Qwen-3.5-0.8B~\cite{qwen3.5}) and VL-SAM~\cite{lin2025vl};
and KnowVal~\citep{xia2025knowval} retrieval and value modules.
We also include KnowVal’s knowledge graph as long-term memory.
The Scheduling Agent is fine-tuned from Qwen3.5-0.8B, requiring $\sim$16 hours on 8 H100 GPUs.

\subsection{Main Results on Driving Benchmarks}
\label{sec:exp:results}

\begin{figure*}[t]
\centering
\begin{subfigure}[t]{0.49\linewidth}
\centering
\includegraphics[width=\linewidth]{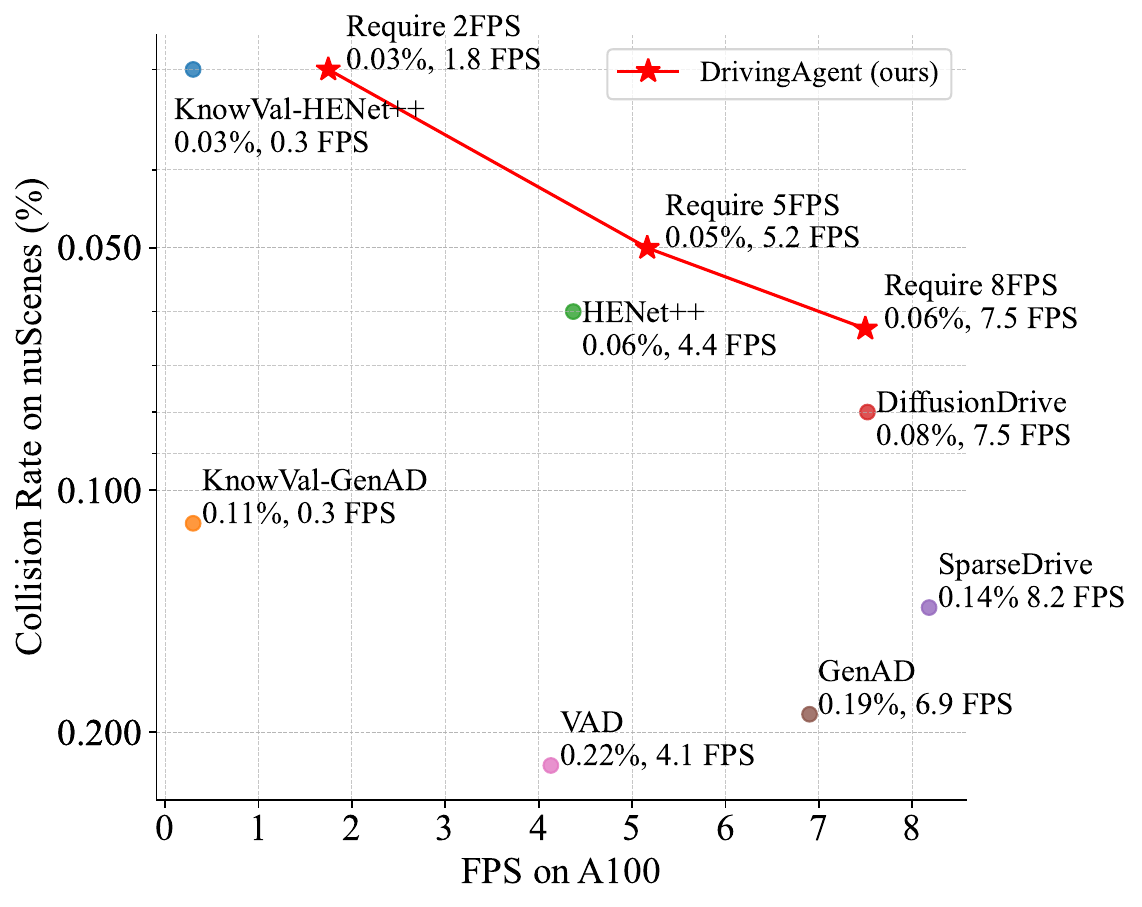}
\caption{\textbf{nuScenes open-loop evaluation.}}
\label{fig:main:nuscenes}
\end{subfigure}
\hfill
\begin{subfigure}[t]{0.49\linewidth}
\centering
\includegraphics[width=\linewidth]{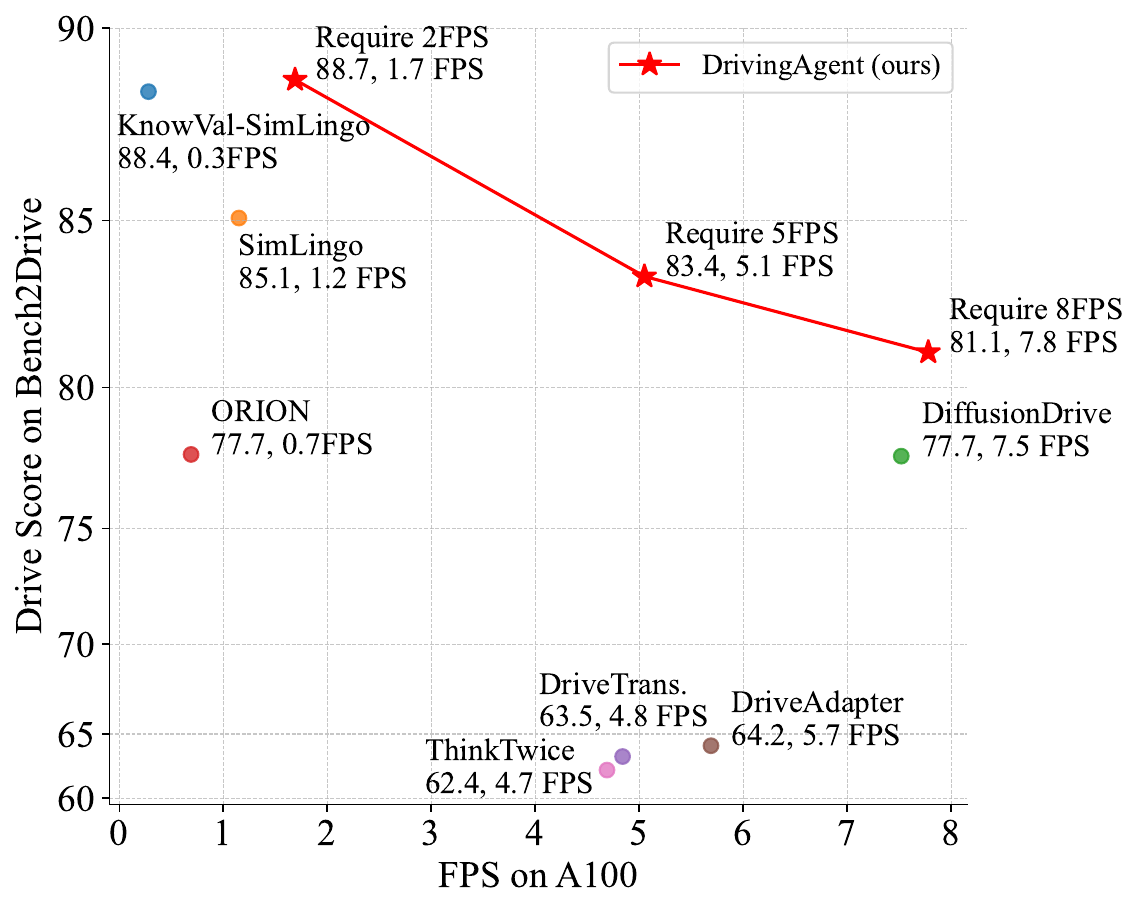}
\caption{\textbf{Bench2Drive closed-loop evaluation.}}
\label{fig:main:b2d}
\end{subfigure}
\caption{\textbf{Speed--accuracy trade-off on nuScenes and Bench2Drive.} Each method is plotted by planning quality versus theoretical compute FPS on an Nvidia A100 GPU. 
DrivingAgent~(ours, \textcolor{red}{$\bigstar$}) is evaluated under three budgets: \emph{Require 2/5/8,FPS}. The scheduler adapts agent-tool invocation without retraining the base planner, achieving a favorable trade-off across both benchmarks.}
\label{fig:main_result}
\vspace{-8pt}
\end{figure*}


As shown in Figure 3, DrivingAgent achieves a better speed–accuracy balance on both the nuScenes and Bench2Drive benchmarks and closely follows the FPS requirements in the instructions.
Detailed tool invocation counts are provided in Table~\ref{tab:appa}.
We also provide an example video of a real driving scenario in the supplementary materials to better illustrate the scheduling agent's working process.
Notably, KnowVal~\cite{xia2025knowval} collects prior designs, then manually assembles them into a system. Our DrivingAgent is given the same prior designs, but not KnowVal's manual system blueprint. DrivingAgent achieves a better speed-accuracy balance by adjusting tool calls and cache reuse across different runtime budgets, demonstrating that it can outperform a manually designed system while using the same component inputs.

Additionally, the proposed parallel scheduling approach may raise a natural concern about whether the FPS would be unstable. We report the coefficient of variation of FPS ($\sigma/\mu$) in Table 3, where a larger value indicates more severe relative fluctuation in FPS. Incorporating GRPO finetuning and the FPS calculator improves FPS stability, reducing $\sigma/\mu$ to a level slightly higher than that of deterministic end-to-end planners, and well below VLA baselines such as ORION and SimLingo, whose autoregressive decoding introduces inherent latency variability.




\subsection{Ablation Study}
\label{sec:exp:ablation}

\textbf{Design Agent effectiveness.}
Initially, only the four HENet++ modules form a connected input‑to‑output pipeline; other capabilities (e.g., VLM-based perception, retrieval) are present but remain disconnected. Directly scheduling these external modules, therefore, leaves the HENet++ computation path unchanged and planning metrics identical to the baseline. The Design Agent bridges this gap by generating adapter modules. As shown in Table~\ref{tab:design_adapter_ablation}, each automatically designed adapter, when combined with the initial modules, reduces the collision rate, directly validating both the adapter and the Design Agent’s effectiveness.




\textbf{Ablation on the Scheduling Agent.}
As shown in Figure~\ref{fig:abl_scheduler} and Table~\ref{tab:full_ablation}, Exp. A uses a zero-shot LLM for scheduling. 
Exp. B fine-tunes the agent framework using GRPO with episodic reward, which simultaneously improves the speed–accuracy balance and FPS stability. 
Exp. C adds a dense reward on top of B, improving accuracy at high FPS. 
Exp. D further incorporates an FPS calculator and feedback as agent tools, thereby enhancing the speed–accuracy balance, improving FPS instruction-following capability, and further reducing FPS fluctuations.

\begin{figure*}[t]
    \centering
    \begin{minipage}[t]{0.54\linewidth}
        \centering
\captionof{table}{\textbf{White-box adapter gains on the HENet++.} 
Each row: original HENet++ modules + one auto-designed adapter. Latency = adapter runtime; Params = adapter parameters.
}
        \label{tab:design_adapter_ablation}
        \vspace{0pt}
        \footnotesize
        \setlength{\tabcolsep}{3pt}
        \resizebox{\linewidth}{!}{
        \begin{tabular}{l|cc|cc}
            \toprule
            AdapterName & Latency & Params & Avg. L2 & Avg. Col \\
            \midrule
            \texttt{\detokenize{Baseline HENet++}} & - & - & 0.647 & 0.057\% \\
            \texttt{\detokenize{IntentGuidedAgentGate}} & 0.48 ms & 82k & 0.630 & 0.031\% \\
            \texttt{\detokenize{IntentConditionedAgentGate}} & 0.91 ms & 363k & 0.599 & 0.053\% \\
            \texttt{\detokenize{FieldConditioned...Bridge}} & 3.03 ms & 341k & 0.745 & 0.049\% \\
            \texttt{\detokenize{ContextSparseAgentGate}} & 0.84 ms & 197k & 0.545 & 0.039\% \\
            \texttt{\detokenize{IntentCreditRouter}} & 0.51 ms & 182k & 0.542 & 0.040\% \\

            ...... &&&&\\
            \bottomrule
        \end{tabular}
        }
    \end{minipage}
    \hfill
    \begin{minipage}[t]{0.44\linewidth}
    \centering
    \captionof{table}{\textbf{FPS Stability.} 
    $R_{\text{epi}}$: GRPO with episodic reward; 
    $R_{\text{dense}}$: $+$ dense reward; 
    Calc: FPS calculator; 
    Para.: paradigm; 
    $\sigma/\mu$: FPS Coefficient of Variation.}
    \label{tab:full_ablation}
    \vspace{0pt}
    \footnotesize
    \setlength{\tabcolsep}{5pt}
    \resizebox{\linewidth}{!}{
    \begin{tabular}{@{}lc|ccc|cc@{}}
        \toprule
        \textbf{Method} & \textbf{ID} & \textbf{\shortstack{$R_{\text{epi}}$}} & \textbf{$R_{\text{dense}}$} & \textbf{Calc} & \textbf{Para.} & \textbf{$\sigma / \mu$} \\
        \midrule
        HENet++         &   & —      & —      & —      & E2E      & 0.037 \\
        DiffusionDrive  &   & —      & —      & —      & E2E      & 0.043 \\
        ORION           &   & —      & —      & —      & VLA      & 0.151 \\
        SimLingo        &   & —      & —      & —      & VLA      & 0.188 \\
        \midrule
        \multirow{4}{*}{DrivingAgent}    & A & \multicolumn{3}{c|}{\textit{zero-shot}} & AgentSys & 0.387 \\
                        & B & \checkmark   & —      & —      & AgentSys & 0.168 \\
                        & C & \checkmark   & \checkmark & —  & AgentSys & 0.172 \\
                        & D & \checkmark   & \checkmark & \checkmark & AgentSys & 0.074 \\
        \bottomrule
    \end{tabular}}
    \end{minipage}
\end{figure*}


\begin{figure*}[t]
\vspace{-5pt}
\centering
\begin{subfigure}[t]{0.49\linewidth}
\centering
\includegraphics[width=\linewidth]{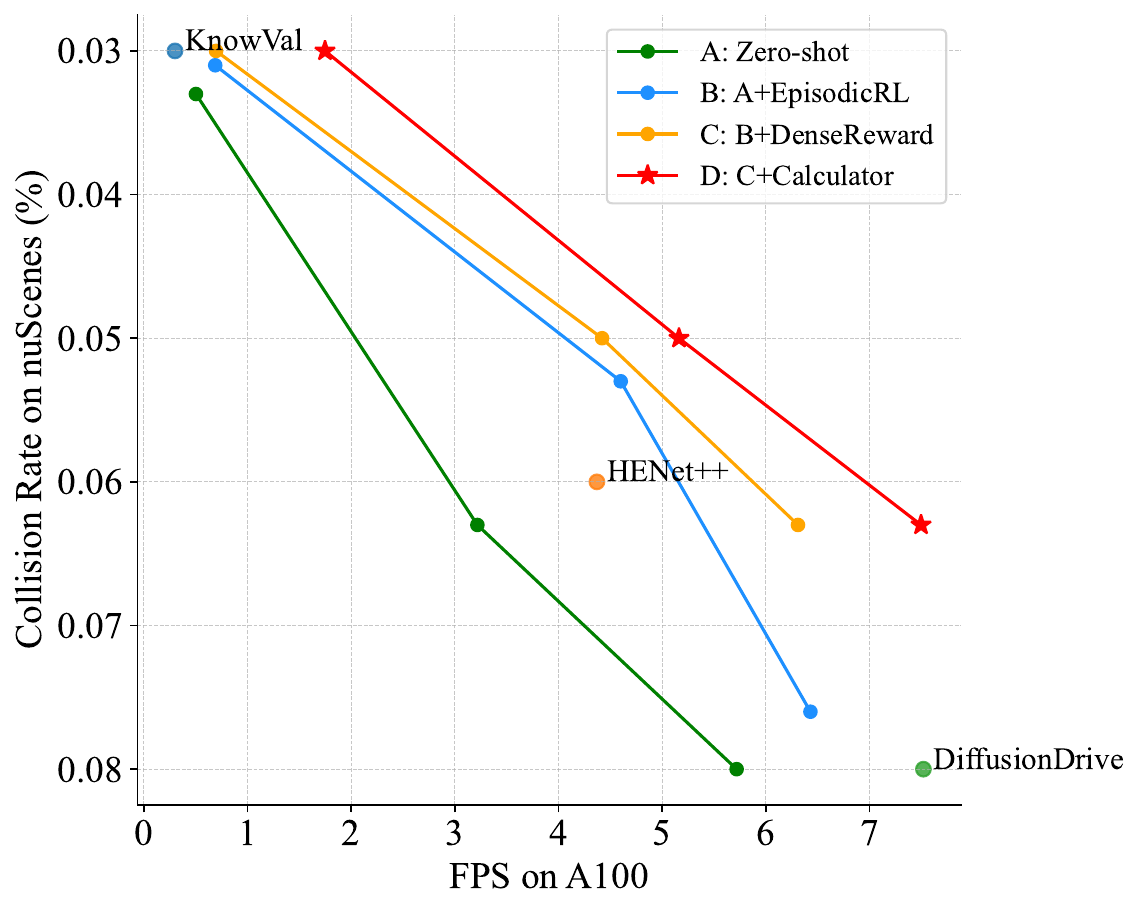}
\label{fig:abl_nus}
\end{subfigure}
\hfill
\begin{subfigure}[t]{0.49\linewidth}
\centering
\includegraphics[width=\linewidth]{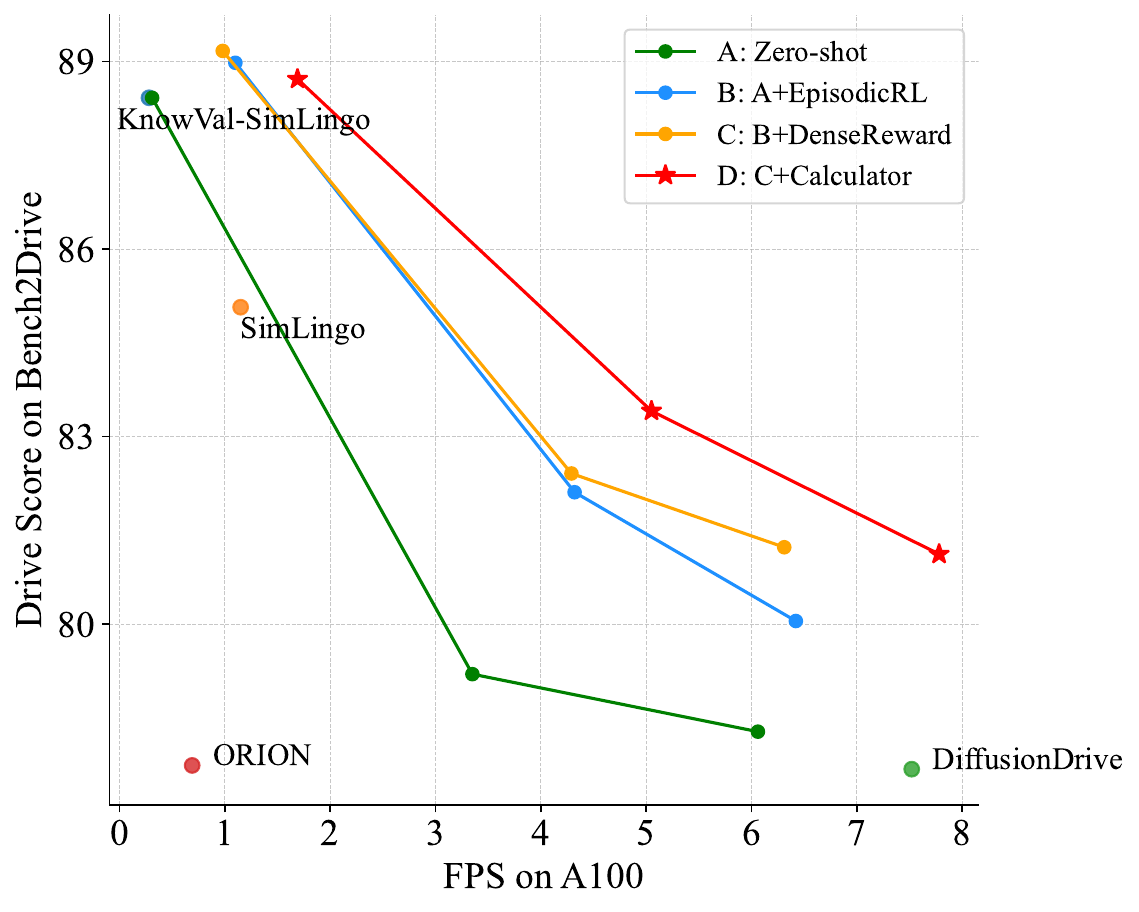}
\label{fig:abl_b2d}
\end{subfigure}
\vspace{-15pt}
\caption{\textbf{Ablation on the Scheduling Agent.}
The experimental results demonstrate the speed–accuracy improvements achieved by GRPO fine-tuning, the proposed reward design, and the addition of an FPS calculator as an agent tool.
}
\label{fig:abl_scheduler}
\vspace{-10pt}
\end{figure*}

%% file: sections/6_conclusion.tex
\section{Conclusion}
\label{sec:conclusion}

We introduced DrivingAgent, a specialized agent framework that unifies the design and scheduling of autonomous driving systems. By treating these as distinct yet coupled tasks, DrivingAgent automates module creation and dynamically orchestrates their execution in real time. Its ability to leverage the internal structure of neural network modules, together with a purpose-built memory system, enables context-aware scheduling that improves resource efficiency. Experiments on nuScenes and Bench2Drive demonstrate that DrivingAgent achieves a favorable speed-accuracy trade-off.

%% file: sections/7_appendix.tex
\section{Experimental Details: Sample Outputs of the Design Agent}

\begin{figure}[h]
  \centering
  \includegraphics[width=\linewidth]{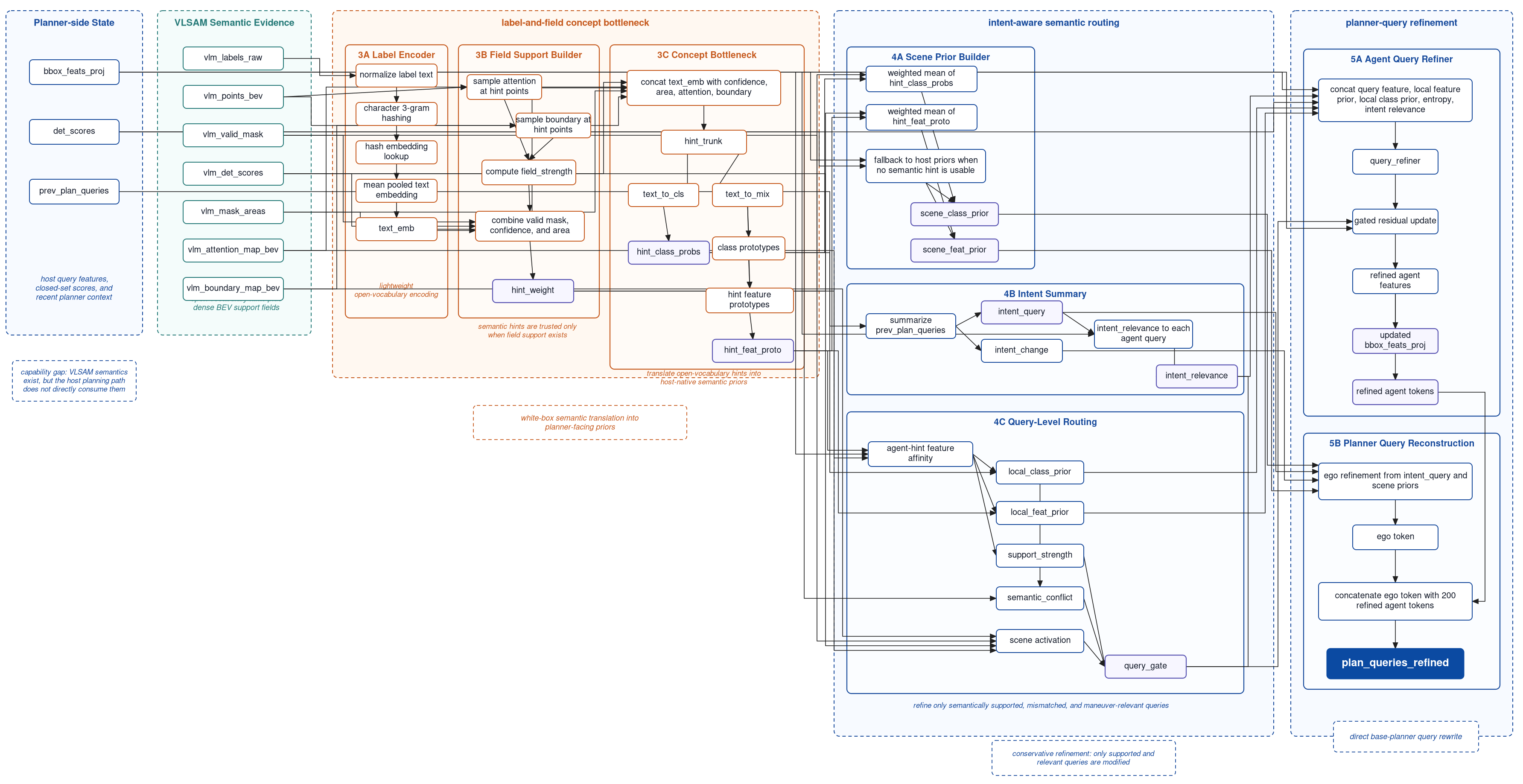}
  \caption{\textbf{FieldConditionedConceptBottleneckBridge.} A white-box VLSAM-to-planning adapter attached at \texttt{pre\_planning}. The module encodes open-vocabulary VLSAM labels, combines them with dense field support, forms a compact concept bottleneck, derives scene-level and query-level semantic priors, and rewrites \texttt{plan\_queries\_refined} for the host planner.}
  \label{fig:appa1}
\end{figure}

\begin{figure}[ht]
  \centering
  \includegraphics[width=\linewidth]{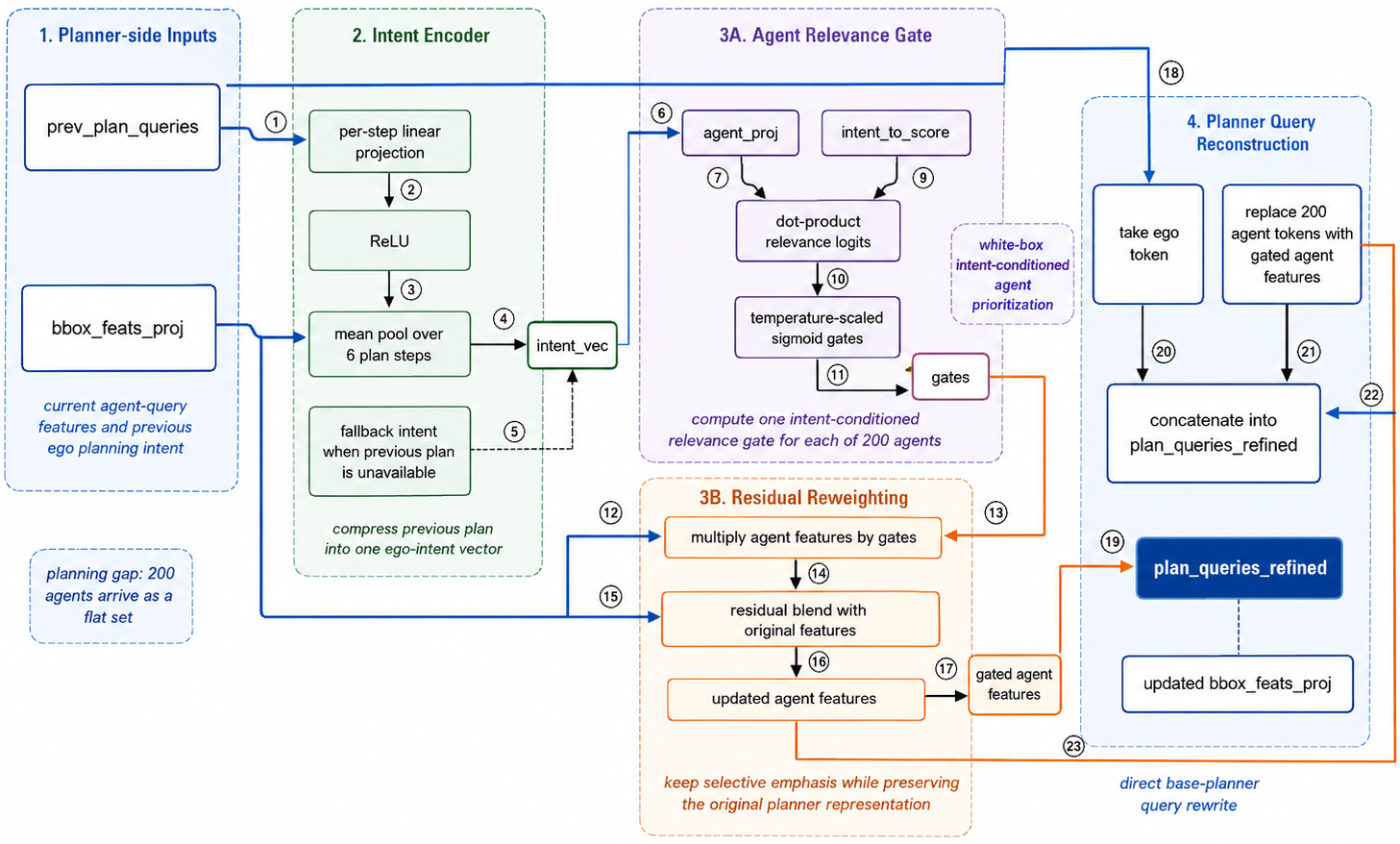}
  \caption{\textbf{IntentGuidedAgentGate.} Architecture of an automatically designed white-box planner-side gating adapter attached at \texttt{pre\_planning}. The module summarizes previous ego intent from \texttt{prev\_plan\_queries}, scores each agent query against that intent, and rewrites \texttt{plan\_queries\_refined} through intent-conditioned agent gating.}
  \label{fig:appa2}
\end{figure}

\begin{figure}[ht]
  \centering
  \includegraphics[width=\linewidth]{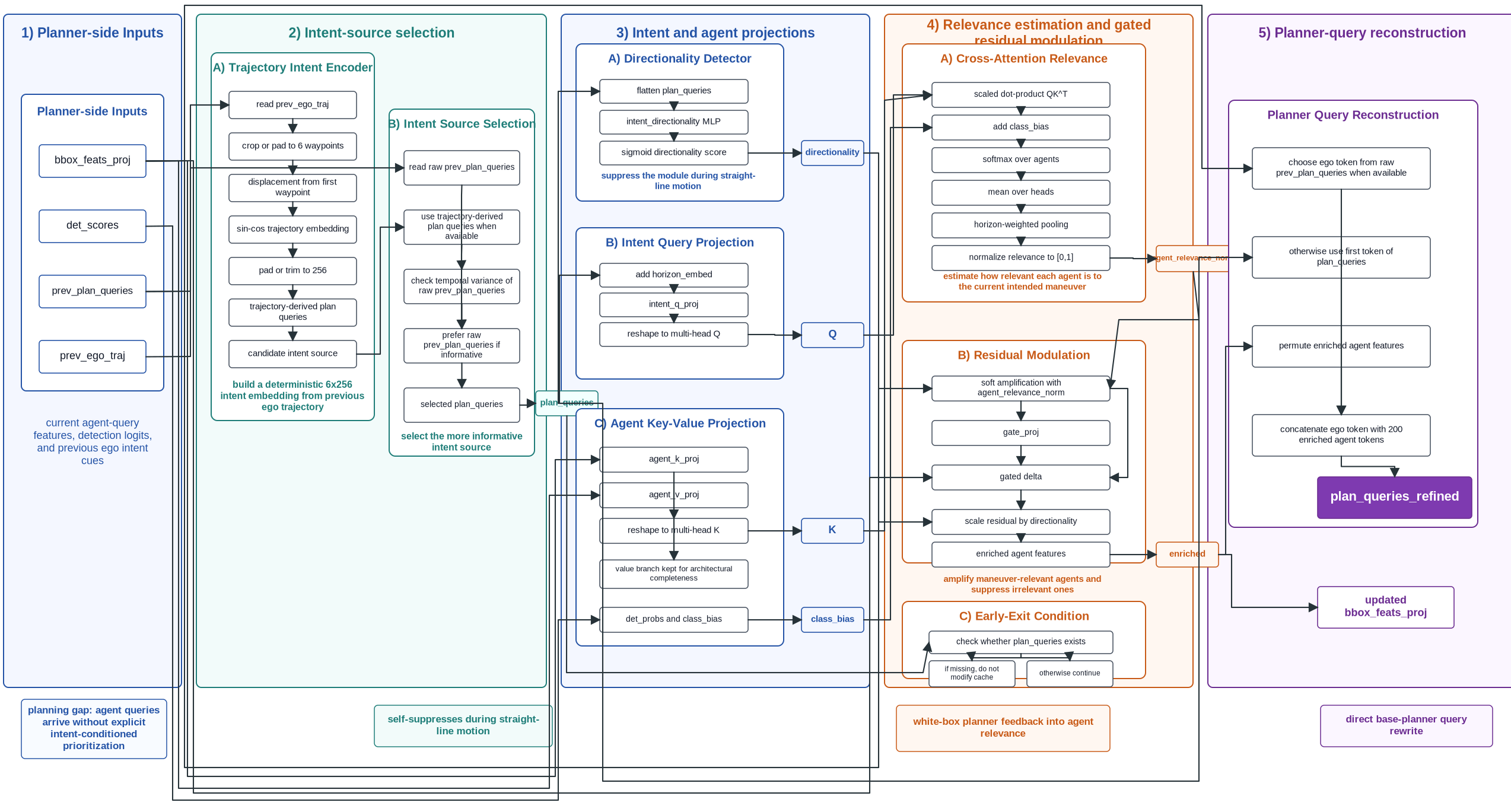}
  \caption{\textbf{IntentConditionedAgentGate.} Architecture of an automatically designed white-box planner-side adapter attached at \texttt{pre\_planning}. The module selects an intent source from trajectory-derived and cached planner context, projects the resulting intent into multi-head query features, attends over planner-side agent tokens, and rewrites \texttt{plan\_queries\_refined} with directionality-aware intent-conditioned agent modulation.}
  \label{fig:appa3}
\end{figure}

\begin{figure}[ht]
  \centering
  \includegraphics[width=\linewidth]{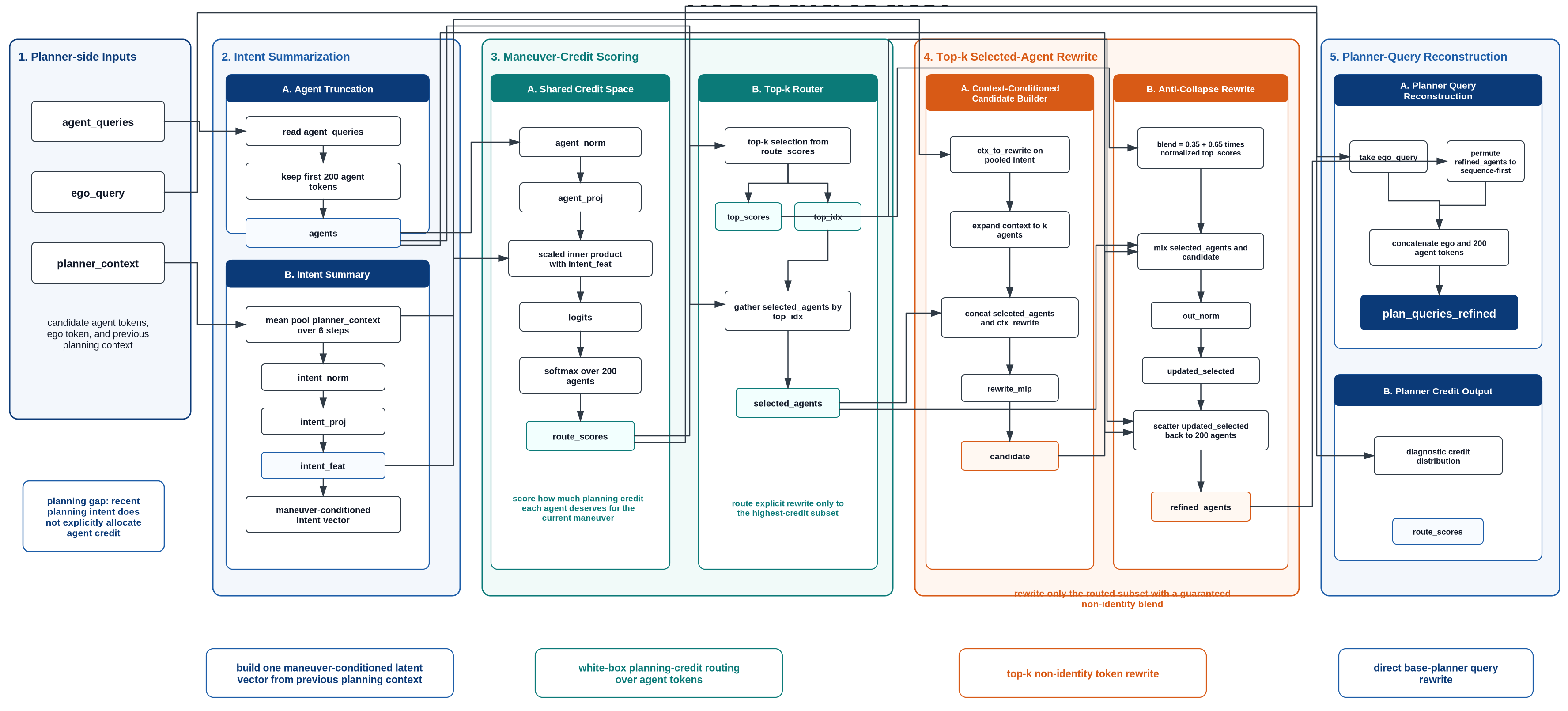}
  \caption{\textbf{IntentCreditRouter.} Architecture of an automatically designed white-box planning router attached at \texttt{pre\_planning}. The module computes per-agent planning credit from previous planner context, selects the highest-credit agent subset, rewrites only those selected tokens with a context-conditioned non-identity update, and reconstructs \texttt{plan\_queries\_refined} for the host planner.}
  \label{fig:appa4}
\end{figure}

Figure~\ref{fig:appa1}--\ref{fig:appa4} shows the specific neural network architectures of several designed modules listed in Table \ref{tab:design_adapter_ablation}.

\clearpage

\section{Experimental Details: Sample of the Profile}

Below is an example illustrating how a profile describes a specific module (Figure~\ref{fig:appa1}).

\begin{lstlisting}[style=profileyaml]
name: FieldConditionedConceptBottleneckBridge
type: adapter_profile
status: appendix_example
host_architecture: HENet++
hook_point: pre_planning
capabilities:
  - vlsam

role: >
  White-box semantic bridge that converts VLSAM hints into planner-facing
  semantic priors and direct query refinement.

purpose: >
  Bridge open-vocabulary VLSAM semantics into the HENet++ planning query path by translating sparse label-plus-field evidence into scene-level and query-level priors, then writing plan_queries_refined.

attachment_rationale: >
  The pre_planning hook is the smallest explicit host attachment space where the module can jointly see planner-side query features, detector-side class scores, recent planner intent, and external VLSAM fields before the planner
  query tensor is consumed downstream.

activation_narrative: >
  Enable this module when valid VLSAM hints are present, local field support is non-trivial, and some planner-side queries remain semantically ambiguous under closed-set detector scores.

inputs:
  required_cache_fields:
    - bbox_feats_proj
    - det_scores
    - vlm_points_bev
    - vlm_valid_mask
    - vlm_det_scores
    - vlm_mask_areas
    - vlm_attention_map_bev
    - vlm_boundary_map_bev
  optional_cache_fields:
    - prev_plan_queries
  required_meta_fields:
    - vlm_labels_raw
  optional_meta_fields:
    - semantic_source
    - vlm_response_raw
    - detseg_boxes_img_raw

outputs:
  main_planner_outputs:
    - bbox_feats_proj
    - plan_queries_refined
  semantic_diagnostics:
    - det_vlm_scene_class_prior
    - det_vlm_scene_feat_prior
    - det_vlm_query_gate
    - det_vlm_activation_score
  training_outputs:
    - det_vlm_text_class_anchor_loss
    - det_vlm_query_class_kl_loss
    - det_vlm_feature_pull_loss
    - det_vlm_improvement_loss
    - det_vlm_bridge_aux_loss

runtime_contract:
  execution_class: inline
  affects_current_frame: true
  latest_wins: false
  output_key: plan_queries_refined
  output_shape: (201, B, 256)

model_profile:
  parameter_count: 341652
  latency_note: >
    Standalone latency should be re-profiled after the current pre_planning
    rewrite.

scheduler_note: >
  This is an inline current-frame module. The scheduling agent should prefer it only on frames where VLSAM semantics are present and semantically relevant to the current maneuver, because the module directly modifies the planner query path.

claim_alignment: >
  This profile supports the paper's white-box claim: external VLSAM semantics are not left as side information, but are injected into the host planner through an explicit attachment point and a typed runtime contract.

main_limitation: >
  Query-to-hint routing is driven mainly by feature-space affinity rather than explicit geometry-aware matching, so the module should be described as structured semantic conditioning rather than exact object-level grounding.
\end{lstlisting}

\section{Experimental Details: Tools Invocation Counts of the Scheduling Agent}

Table~\ref{tab:appa} below reports the average latency and invocation count of each module (including the scheduler LLM). The six columns correspond to Zero‑Shot (A) and the full DrivingAgent (D) in Figure 4, each evaluated under three different FPS requirements.

\begin{table}[h]
    \centering
    \caption{The average latency and invocation count of each module.}
    \label{tab:appa}
    \resizebox{\linewidth}{!}{
    \begin{tabular}{l|r|ccc|ccc}
    \toprule
    \multirow{2}{*}{Tool name}  & {Latency} & \multicolumn{3}{c|}{DrivingAgent Zero-Shot} & \multicolumn{3}{c}{DrivingAgent GRPO}\\
     & {Avg. (ms)} & 2 FPS & 5 FPS & 8 FPS & 2 FPS & 5 FPS & 8 FPS \\
    \midrule
    \gr Scheduling-Agent &651.45& 1744 & 845 & 315 & 1029 & 429 & 208 \\
    FPS Calculator &0.00& 0 & 0 & 0 & 1029 & 429 & 208 \\
    \gr HENet-Backbone &11.85& 6019 & 6019 & 6019 & 6019 & 6019 & 0 \\
    HENet-Detection &41.85& 6019 & 2762 & 229 & 5981 & 3780 & 0 \\
    \gr HENet-Occupancy &164.43& 6019 & 2528 & 229 & 5019 & 628 & 0 \\
    HENet-Planner &10.63& 6019 & 6019 & 6019 & 6019 & 6019 & 0 \\
    \gr DiffDrive-Backbone &12.93& 0 & 0 & 0 & 0 & 0 & 6019 \\
    DiffDrive-Perception &76.89& 0 & 0 & 0 & 0 & 0 & 2438 \\
    \gr DiffDrive-Planner &33.34& 0 & 0 & 0 & 0 & 0 & 6019 \\
    Knowledge Retrieval &82.70& 2008 & 412 & 432 & 617 & 405 & 43 \\
    \gr VLM-0.8B &446.44& 6019 & 390 & 396 & 1461 & 341 & 225 \\
    VL-SAM Adapter + SAM &1095.89& 6019 & 390 & 396 & 753 & 249 & 64 \\
    \gr IntentGuidedAgentGate &0.48& 6019 & 3036 & 598 & 6003 & 5152 & 5006 \\
    IntentConditionedAgentGate &0.91& 6019 & 3030 & 598 & 5991 & 4980 & 3782 \\
    \gr FieldConditioned...Bridge &3.03& 6019 & 3030 & 591 & 6019 & 6009 & 6018 \\
    ContextSparseAgentGate &0.84& 6019 & 3036 & 598 & 6003 & 5152 & 5006 \\
    \gr IntentCreditRouter &0.51& 6019 & 3030 & 597 & 5998 & 5399 & 5084 \\
    ... &&&&&&&\\
    \midrule
    \multicolumn{2}{l|}{Total Frames} & 6019 & 6019 & 6019 & 6019 & 6019 & 6019 \\
    \gr \multicolumn{2}{l|}{Avg. FPS} & 0.502 & 3.217 & 5.718 & 1.747 & 5.162 & 7.497 \\
    \midrule
    \multirow{4}{*}{Collision Rate} & 1s & 0.00 & 0.00 & 0.00 & 0.00 & 0.00 & 0.00 \\
    & 2s & 0.00 & 0.01 & 0.00 & 0.01 & 0.01 & 0.04 \\
    & 3s & 0.10 & 0.18 & 0.24 & 0.08 & 0.14 & 0.15 \\
    \gr & Avg. & 0.03 & 0.06 & 0.08 & 0.03 & 0.05 & 0.06 \\

    \bottomrule
    \end{tabular}}
    \vspace{1mm}
    
\end{table}

The scheduling results above indicate that when the scheduling agent employs a GRPO fine-tuned LLM, it achieves a better speed–accuracy balance. 
For instance, between the two perception tools \textit{HENet-Detection} and \textit{HENet-Occupancy}, the scheduling agent learns to reduce calls to the slower \textit{HENet-Occupancy} while maintaining high-frequency scheduling of the lightweight adapter tools produced by the design agent. 
Furthermore, the scheduling agent spontaneously learns to decouple the VLM of VL-SAM from its subsequent modules: when the VLM detects no long-tail objects, it no longer invokes the downstream VL-SAM adapter and SAM. 
Moreover, the VLM is no longer limited to serving VL-SAM. Its retained features are transformed by other adapter tools designed by the design agent and reused in other modules.



%% file: references.bib
@String(CVPR= {IEEE Conf. Comput. Vis. Pattern Recog.})

@String(ICCV= {Int. Conf. Comput. Vis.})

@String(ECCV= {Eur. Conf. Comput. Vis.})

@String(NIPS= {Adv. Neural Inform. Process. Syst.})

@String(ICLR = {Int. Conf. Learn. Represent.})

@String(CVPR  = {CVPR})

@String(ICCV  = {ICCV})

@String(ECCV  = {ECCV})

@String(NIPS  = {NeurIPS})

@String(ICLR  = {ICLR})

@String(ICRA = {ICRA})

@String(COLM = {COLM})

@String(UIST = {UIST})

@inproceedings{hu2023uniad,  
title={Planning-oriented Autonomous Driving}, 
author={Hu, Yihan and Yang, Jiazhi and Chen, Li and Li, Keyu and Sima, Chonghao and Zhu, Xizhou and Chai, Siqi and Du, Senyao and Lin, Tianwei and Wang, Wenhai and Lu, Lewei and Jia, Xiaosong and Liu, Qiang and Dai, Jifeng and Qiao, Yu and Li, Hongyang}, 
booktitle=CVPR,
year={2023}, 
}

@inproceedings{caesar2020nuscenes,  
title={nuScenes: A multimodal dataset for autonomous driving}, 
booktitle=CVPR, 
author={Caesar, Holger and Bankiti, Varun and Lang, Alex H. and Vora, Sourabh and Liong, Venice Erin and Xu, Qiang and Krishnan, Anush and Pan, Yu and Baldan, Giancarlo and Beijbom, Oscar},  
year={2020}, 
}

@inproceedings{hu2022st,
  title={St-p3: End-to-end vision-based autonomous driving via spatial-temporal feature learning},
  author={Hu, Shengchao and Chen, Li and Wu, Penghao and Li, Hongyang and Yan, Junchi and Tao, Dacheng},
  booktitle=ECCV,
  year={2022},
}

@inproceedings{jiang2023vad,
  title={Vad: Vectorized scene representation for efficient autonomous driving},
  author={Jiang, Bo and Chen, Shaoyu and Xu, Qing and Liao, Bencheng and Chen, Jiajie and Zhou, Helong and Zhang, Qian and Liu, Wenyu and Huang, Chang and Wang, Xinggang},
  booktitle=ICCV,
  year={2023}
}

@inproceedings{weng2024drive,
  title={Para-drive: Parallelized architecture for real-time autonomous driving},
  author={Weng, Xinshuo and Ivanovic, Boris and Wang, Yan and Wang, Yue and Pavone, Marco},
  booktitle=CVPR,
  year={2024}
}

@inproceedings{zheng2024genad,
  title={Genad: Generative end-to-end autonomous driving},
  author={Zheng, Wenzhao and Song, Ruiqi and Guo, Xianda and Zhang, Chenming and Chen, Long},
  booktitle=ECCV,
  year={2024},
}

@inproceedings{song2025don,
  title={Don't Shake the Wheel: Momentum-Aware Planning in End-to-End Autonomous Driving},
  author={Song, Ziying and Jia, Caiyan and Liu, Lin and Pan, Hongyu and Zhang, Yongchang and Wang, Junming and Zhang, Xingyu and Xu, Shaoqing and Yang, Lei and Luo, Yadan},
  booktitle=CVPR,
  year={2025}
}

@inproceedings{zhang2025bridging,
  title={Bridging past and future: End-to-end autonomous driving with historical prediction and planning},
  author={Zhang, Bozhou and Song, Nan and Jin, Xin and Zhang, Li},
  booktitle=CVPR,
  year={2025}
}

@inproceedings{sun2025sparsedrive,
  title={Sparsedrive: End-to-end autonomous driving via sparse scene representation},
  author={Sun, Wenchao and Lin, Xuewu and Shi, Yining and Zhang, Chuang and Wu, Haoran and Zheng, Sifa},
  booktitle=ICRA,
  year={2025},
}

@inproceedings{renz2025simlingo,
  title={Simlingo: Vision-only closed-loop autonomous driving with language-action alignment},
  author={Renz, Katrin and Chen, Long and Arani, Elahe and Sinavski, Oleg},
  booktitle=CVPR,
  year={2025}
}

@inproceedings{wu2022trajectory,
  title={Trajectory-guided control prediction for end-to-end autonomous driving: A simple yet strong baseline},
  author={Wu, Penghao and Jia, Xiaosong and Chen, Li and Yan, Junchi and Li, Hongyang and Qiao, Yu},
  booktitle=NIPS,
  year={2022}
}

@inproceedings{jia2023think,
  title={Think twice before driving: Towards scalable decoders for end-to-end autonomous driving},
  author={Jia, Xiaosong and Wu, Penghao and Chen, Li and Xie, Jiangwei and He, Conghui and Yan, Junchi and Li, Hongyang},
  booktitle=CVPR,
  year={2023}
}

@inproceedings{jia2023driveadapter,
  title={Driveadapter: Breaking the coupling barrier of perception and planning in end-to-end autonomous driving},
  author={Jia, Xiaosong and Gao, Yulu and Chen, Li and Yan, Junchi and Liu, Patrick Langechuan and Li, Hongyang},
  booktitle=ICCV,
  year={2023}
}

@inproceedings{jia2024bench2drive,
  title={Bench2drive: Towards multi-ability benchmarking of closed-loop end-to-end autonomous driving},
  author={Jia, Xiaosong and Yang, Zhenjie and Li, Qifeng and Zhang, Zhiyuan and Yan, Junchi},
  booktitle=NIPS,
  year={2024}
}

@inproceedings{liao2025diffusiondrive,
  title={Diffusiondrive: Truncated diffusion model for end-to-end autonomous driving},
  author={Liao, Bencheng and Chen, Shaoyu and Yin, Haoran and Jiang, Bo and Wang, Cheng and Yan, Sixu and Zhang, Xinbang and Li, Xiangyu and Zhang, Ying and Zhang, Qian and others},
  booktitle=CVPR,
  year={2025}
}

@inproceedings{jia2025drivetransformer,
  title={Drivetransformer: Unified transformer for scalable end-to-end autonomous driving},
  author={Jia, Xiaosong and You, Junqi and Zhang, Zhiyuan and Yan, Junchi},
  booktitle=ICLR,
  year={2025}
}

@inproceedings{fu2025orion,
  title={Orion: A holistic end-to-end autonomous driving framework by vision-language instructed action generation},
  author={Fu, Haoyu and Zhang, Diankun and Zhao, Zongchuang and Cui, Jianfeng and Liang, Dingkang and Zhang, Chong and Zhang, Dingyuan and Xie, Hongwei and Wang, Bing and Bai, Xiang},
  booktitle=ICCV,
  year={2025}
}

@article{henetpp,
  title={HENet++: Hybrid Encoding and Multi-task Learning for 3D Perception and End-to-end Autonomous Driving},
  author={Zhongyu Xia and Zhiwei Lin and Yongtao Wang and Ming-Hsuan Yang},
  journal={arXiv preprint arXiv:2511.07106},
  year={2025},
}

@article{lin2025vl,
  title={VL-SAM-V2: Open-World Object Detection with General and Specific Query Fusion},
  author={Lin, Zhiwei and Wang, Yongtao},
  journal={arXiv preprint arXiv:2505.18986},
  year={2025}
}

@inproceedings{sima2024drivelm,
  title={DriveLM: Driving with Graph Visual Question Answering},
  author={Chonghao Sima and Katrin Renz and Kashyap Chitta and Li Chen and Hanxue Zhang and Chengen Xie and Jens Beißwenger and Ping Luo and Andreas Geiger and Hongyang Li},
  booktitle=ECCV,
  year={2024},
}

@article{han2025percept,
  title={Percept-WAM: Perception-enhanced world-awareness-action model for robust end-to-end autonomous driving},
  author={Han, Jianhua and Tian, Meng and Zhu, Jiangtong and He, Fan and Zhang, Huixin and Guo, Sitong and Zhu, Dechang and Tang, Hao and Xu, Pei and Guo, Yuze and others},
  journal={arXiv preprint arXiv:2511.19221},
  year={2025}
}

@article{xia2025knowval,
  title={KnowVal: A Knowledge-Augmented and Value-Guided Autonomous Driving System},
  author={Xia, Zhongyu and Chen, Wenhao and Wang, Yongtao and Yang, Ming-Hsuan},
  journal={arXiv preprint arXiv:2512.20299},
  year={2025}
}

@article{tian2024drivevlm,
  title={Drivevlm: The convergence of autonomous driving and large vision-language models},
  author={Tian, Xiaoyu and Gu, Junru and Li, Bailin and Liu, Yicheng and Wang, Yang and Zhao, Zhiyong and Zhan, Kun and Jia, Peng and Lang, Xianpeng and Zhao, Hang},
  journal={arXiv preprint arXiv:2402.12289},
  year={2024}
}

@article{li2026unidrivevla,
  title={UniDriveVLA: Unifying Understanding, Perception, and Action Planning for Autonomous Driving},
  author={Li, Yongkang and Zhou, Lijun and Yan, Sixu and Liao, Bencheng and Yan, Tianyi and Xiong, Kaixin and Chen, Long and Xie, Hongwei and Wang, Bing and Chen, Guang and others},
  journal={arXiv preprint arXiv:2604.02190},
  year={2026}
}

@inproceedings{yao2022react,
  title={React: Synergizing reasoning and acting in language models},
  author={Yao, Shunyu and Zhao, Jeffrey and Yu, Dian and Du, Nan and Shafran, Izhak and Narasimhan, Karthik R and Cao, Yuan},
  booktitle=ICLR,
  year={2022}
}

@article{shao2024deepseekmath,
  title={Deepseekmath: Pushing the limits of mathematical reasoning in open language models},
  author={Shao, Zhihong and Wang, Peiyi and Zhu, Qihao and Xu, Runxin and Song, Junxiao and Bi, Xiao and Zhang, Haowei and Zhang, Mingchuan and Li, YK and Wu, Yang and others},
  journal={arXiv preprint arXiv:2402.03300},
  year={2024}
}

@misc{qwen3.5,
    title  = {{Qwen3.5}: Towards Native Multimodal Agents},
    author = {{Qwen Team}},
    month  = {February},
    year   = {2026},
    url    = {https://qwen.ai/blog?id=qwen3.5}
}

@inproceedings{shinn2023reflexion,
  title={Reflexion: Language agents with verbal reinforcement learning},
  author={Shinn, Noah and Cassano, Federico and Gopinath, Ashwin and Narasimhan, Karthik and Yao, Shunyu},
  booktitle=NIPS,
  year={2023}
}

@article{wang2023voyager,
  title={Voyager: An open-ended embodied agent with large language models},
  author={Wang, Guanzhi and Xie, Yuqi and Jiang, Yunfan and Mandlekar, Ajay and Xiao, Chaowei and Zhu, Yuke and Fan, Linxi and Anandkumar, Anima},
  journal={arXiv preprint arXiv:2305.16291},
  year={2023}
}

@inproceedings{wu2024autogen,
  title={Autogen: Enabling next-gen LLM applications via multi-agent conversations},
  author={Wu, Qingyun and Bansal, Gagan and Zhang, Jieyu and Wu, Yiran and Li, Beibin and Zhu, Erkang and Jiang, Li and Zhang, Xiaoyun and Zhang, Shaokun and Liu, Jiale and others},
  booktitle=COLM,
  year={2024}
}

@inproceedings{hong2023metagpt,
  title={MetaGPT: Meta programming for a multi-agent collaborative framework},
  author={Hong, Sirui and Zhuge, Mingchen and Chen, Jonathan and Zheng, Xiawu and Cheng, Yuheng and Wang, Jinlin and Zhang, Ceyao and Wang, Zili and Yau, Steven Ka Shing and Lin, Zijuan and others},
  booktitle=ICLR,
  year={2023}
}

@inproceedings{park2023generative,
  title={Generative agents: Interactive simulacra of human behavior},
  author={Park, Joon Sung and O'Brien, Joseph and Cai, Carrie Jun and Morris, Meredith Ringel and Liang, Percy and Bernstein, Michael S},
  booktitle=UIST,
  year={2023}
}

@inproceedings{schick2023toolformer,
  title={Toolformer: Language models can teach themselves to use tools},
  author={Schick, Timo and Dwivedi-Yu, Jane and Dess{\`\i}, Roberto and Raileanu, Roberta and Lomeli, Maria and Hambro, Eric and Zettlemoyer, Luke and Cancedda, Nicola and Scialom, Thomas},
  booktitle=NIPS,
  year={2023}
}

@inproceedings{chen2024asynchronous,
  title={Asynchronous large language model enhanced planner for autonomous driving},
  author={Chen, Yuan and Ding, Zi-han and Wang, Ziqin and Wang, Yan and Zhang, Lijun and Liu, Si},
  booktitle=ECCV,
  year={2024},
}

@inproceedings{chen2025solve,
  title={Solve: Synergy of language-vision and end-to-end networks for autonomous driving},
  author={Chen, Xuesong and Huang, Linjiang and Ma, Tao and Fang, Rongyao and Shi, Shaoshuai and Li, Hongsheng},
  booktitle=CVPR,
  year={2025}
}

@inproceedings{zhengdriveagent,
  title={DriveAgent-R1: Advancing VLM-based Autonomous Driving with Active Perception and Hybrid Thinking},
  author={Zheng, Weicheng and Mao, Xiaofei and Ye, Nanfei and Li, Pengxiang and Zhan, Kun and Lang, XianPeng and Zhao, Hang},
  booktitle=ICLR,
  year={2026}
}
